\documentclass[10pt,twocolumn,letterpaper]{article}

\usepackage{iccv}
\usepackage{times}
\usepackage{epsfig}
\usepackage{graphicx}
\usepackage{amsmath}
\usepackage{amssymb}
\usepackage{booktabs,cuted}
\usepackage{xcolor}

\usepackage[breaklinks=true,letterpaper=true,colorlinks,bookmarks=false]{hyperref}

\iccvfinalcopy 

\def\iccvPaperID{****} 

\ificcvfinal\fi
\begin{document}

\title{Multi-kernel learning of deep convolutional features for action recognition}

\author{Biswa Sengupta\\
Imperial College London\\
Noah's Ark Lab (Huawei Technologies UK)\\
{\tt\small b.sengupta@imperial.ac.uk}
\and
Yu Qian\thanks{joint first author}\\
Cortexica Vision Systems Limited\\
}

\maketitle

\begin{abstract}
Image understanding using deep convolutional network has reached human-level performance, yet a closely related problem of video understanding especially, action recognition has not reached the requisite level of maturity. We combine multi-kernels based support-vector-machines (SVM) with a multi-stream deep convolutional neural network to achieve close to state-of-the-art performance on a 51-class  activity recognition problem (HMDB-51 dataset); this specific dataset has proved to be particularly challenging for deep neural networks due to the heterogeneity in camera viewpoints, video quality, etc.  The resulting architecture is named pillar networks as each (very) deep neural network acts as a pillar for the hierarchical classifiers. In addition, we illustrate that hand-crafted features such as improved dense trajectories (iDT) and Multi-skip Feature Stacking (MIFS), as additional pillars, can further supplement the performance.

\end{abstract}

\section{Introduction}

Video understanding is a computer vision problem that has attracted the deep-learning community, notably via the usage of the two-stream convolutional network \cite{Simonyan2014}. Such a framework uses a deep convolutional neural network (dCNN) to extract static RGB (Red-Green-Blue) features as well as motion cues from another network that deconstructs the optic-flow of a given video clip. Notably, there has been plenty of work in utilising different types of network architectures for factorising the RGB and optical-flow based features. For example, an inception network \cite{Szegedy2016} uses $1 \times 1$ convolutions in its inception block to estimate cross-channel corrections, which is then followed by the estimation of cross-spatial and cross-channel correlations. A residual network (ResNet), on the other hand, learns residuals on the inputs \cite{He2016}.

There are obvious problems that have impeded high accuracy of deep neural networks for video classification. Videos unlike still images have short and long temporal correlations, attributes that single frame convolutional neural network fail to discover. Therefore, the first hurdle is designing recurrent networks and feedforward networks that can learn this latent temporal structure. Nonetheless, there has been much progress in devising novel neural network architecture since the work of  \cite{Karpathy2014}. Another problem is the large storage and memory requirement for analysing moderately sized video snippets. One requires a relatively larger computing resource to train ultra deep neural networks that can learn the subtleties in temporal correlations, given varying lighting, camera angles, pose, etc. It is also difficult to utilise classical image augmentation techniques on a video stream. Additionally, video-based features (unlike in static images) evolve with a dynamics across several orders of time-scales. To add to this long list of technical difficulties, is the problem of the semantic gap, i.e., whether classification/labelling/captioning can lead to ``understanding" the video snippet?

We improve upon existing technology by combining Inception networks and ResNets using a Support-Vector-Machine (SVM) classifier that is further combined in a multi-kernel setting to yield, to the best of our knowledge, an increased performance on the HMDB51 data-set \cite{Kuehne2013}. Notably, our work makes the following contributions:

\begin{itemize}
\item We introduce pillar networks that are deep as well as wide (depending on use-case), enabling horizontal scalability. This is important for a production quality video analytics framework that has to operate under the constraints of computational time. 
\item HMDB-51 is a dataset that has a wide variety of heterogeneity -- camera angle, video quality, pose, etc. Given that our framework has higher accuracy on this dataset, our methodology will be applicable to datasets that have similar statistical heterogeneity embedded in them.
\end{itemize}

\section{Methods}

In this section, we describe the dataset, the network architectures and the multi-kernel learning based support-vector-machine (SVM) setup that we utilise in our four-stream dCNN pillar network for activity recognition. We refer the readers to the original network architectures in \cite{Wang2016} and \cite{Ma2017} for further technical details. While we do not report the results here,  classification methodologies like AdaBoost, gradient boosting, random forests, etc. have classification accuracy in the range of 5-55\% for this dataset, for either the RGB or the optic-flow based features.

\subsection{Dataset}
The HMDB51 dataset  \cite{Kuehne2013} is an action classification dataset that comprises of 6,766 video clips which have been divided into 51 action classes.   Although a much larger UCF-sports dataset exists with 101 action classes \cite{Soomro2012}, the HMDB51 has proven to be more challenging.  This is because each video has been filmed using a variety of viewpoints, occlusions, camera motions, video quality, etc. anointing the challenges of video-based prediction problems. The second motivation behind using such a dataset lies in the fact that HMDB51 has storage and compute requirement that is fulfilled by a modern workstation with GPUs --   alleviating deployment on expensive cloud-based compute resources. 

All experiments were done on Intel Xeon E5-2687W 3 GHz 128 GB workstation with two 12GB nVIDIA TITAN Xp GPUs. As in the original evaluation scheme, we report accuracy as an average over the three training/testing splits.

\begin{figure*}[h]
\centering
\includegraphics[scale=0.15]{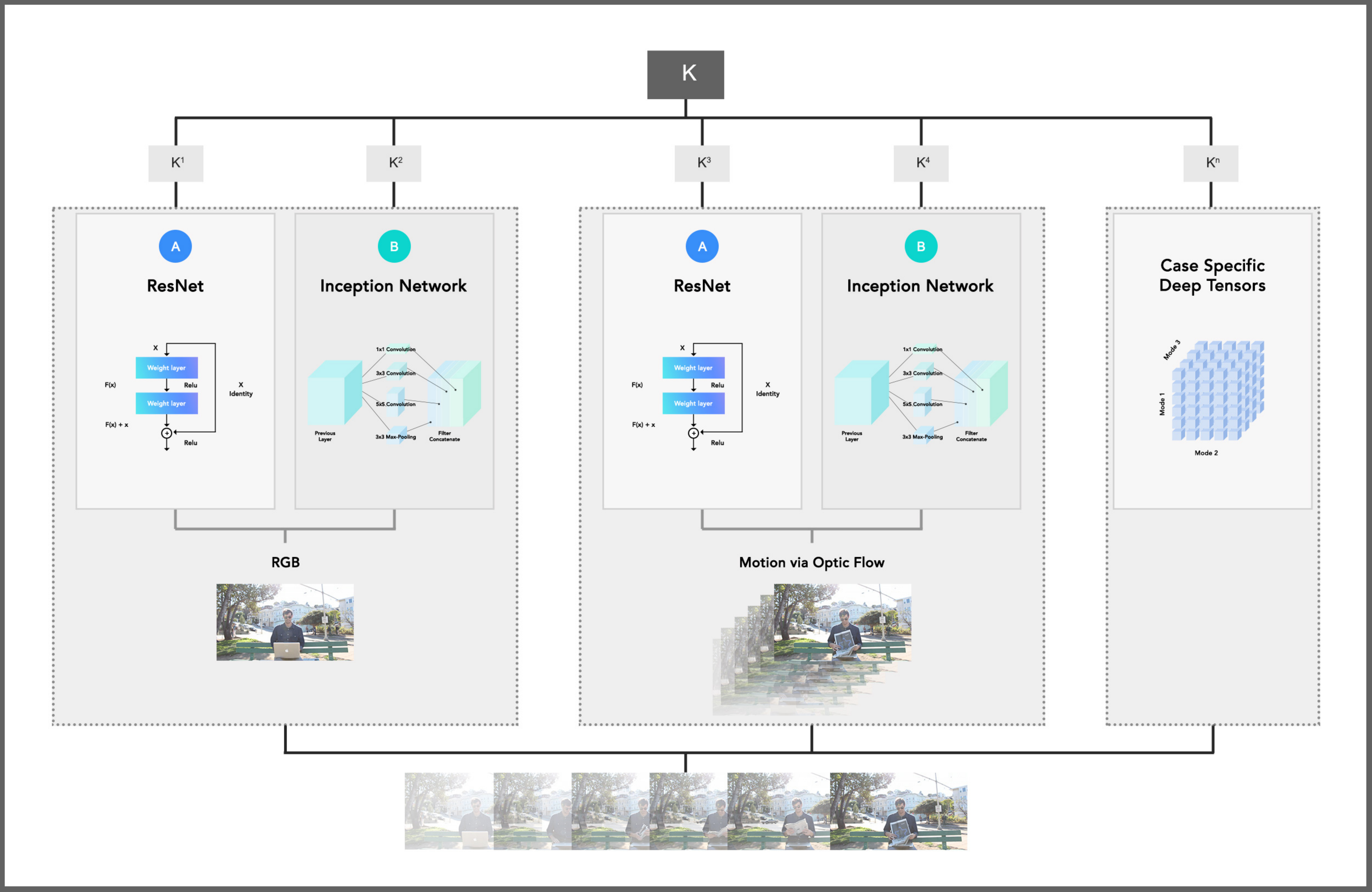}
\caption{\textbf{The Pillar Network framework: }In this specific instantiation, there are two types of networks, namely ResNets and Inception networks that factorise static (RGB) and dynamic (optic flow) inputs obtained from a video. For the case specific deep tensors we use iDT and MIFS, under a multi-kernel learning framework. Additional feature tensors (hand-crafted or otherwise) can be learnt, according to the specific need of the problem, and incorporated as a new pillar.}
 \label{fig:framework}
\end{figure*}

\subsection{Inception layers for RGB and flow extraction}
We use the inception layer architecture described in \cite{Wang2016}. Each video is divided into $N$ segments, and a short sub-segment is randomly selected from each segment so that a preliminary prediction can be produced from each snippet. This is later combined to form a video-level prediction. An Inception with Batch Normalisation network \cite{Ioffe2015} is utilised for both the spatial and the optic-flow stream.  The feature size of each inception network is fixed at 1024. For further details on network pre-training, construction, etc. please refer to \cite{Wang2016}.

\subsection{Residual layers for RGB and flow extraction}
We utilise the network architecture proposed in \cite{Ma2017} where the authors leverage recurrent networks and convolutions over temporally constructed feature matrices as shown in Fig. \ref{fig:framework}. In our instantiation, we truncate the network to yield 2048 features, which is different from \cite{Ma2017} where these features feed into an LSTM (Long Short Term Memory) network. The spatial stream network takes in RGB images as input with a ResNet-101 \cite{He2016} as a feature extractor; this ResNet-101 spatial-stream ConvNet has been pre-trained on the ImageNet dataset. The temporal stream stacks ten optical flow images using the pre-training protocol suggested in \cite{Wang2016}.  The feature size of each ResNet network is fixed at 2048. For further details on network pre-training, construction, etc. please refer to \cite{Ma2017}.

\subsection{Hand-crafted features: iDT and MIFS}

We follow \cite{Wang2013a}  for generating the Fisher encoded Improved Dense Trajectory (iDT) features\footnote{http://lear.inrialpes.fr/~wang/improved\_trajectories}. First tracking points are created by median filtering in a dense optical flow field of 15 frames. We can then compute descriptors such as Trajectory, a histogram of oriented gradients (HOG), a histogram of optical flow (HOF) and motion boundary histograms (MBH)   \cite{Wang2013a}. Descriptors within a space-time volume are aligned with a trajectory to encode the motion information; Fisher encoding \cite{Perronnin2007} is then applied on the local descriptors to generate the video representation for classification. 

Similar to Fisher encoded iDT, instead of using feature points extracted from one-time scale, Multi-skIp Feature Stacking (MIFS) \cite{Lan2015} extracts and stacks all of the raw feature points from multiple time skips (scale) before encoding it to a Fisher vector. MIFS achieves shift-invariance in the frequency domain and captures a longer range of action indicators by recapturing information at coarse scales.

\subsection{Support Vector Machine (SVM) with multi-kernel learning (MKL)}

The basis of the second stage of our classification methodology rests on a maximum margin classifier -- a support vector machine (SVM). Given training tuples $(x_{i},y_{i})$ and weights $w$ , under a Hinge loss, a SVM solves the primal problem \cite{Scholkopf2001},

\begin{eqnarray}
  \mathop {\min }\limits_{w,b,\zeta } \frac{1}{2}{w^T}w & + & C\sum\limits_{i = 1}^n {{\zeta _i}}  \nonumber \\
&  s.t. & {y_i}\left( {{w^T}\phi \left( {{x_i}} \right) + b} \right) \geqslant 1 - {\zeta _i} \nonumber \\
& & {\zeta _i} \geqslant 0,i = 1, \ldots ,n \nonumber \\
\end{eqnarray}

As is customary in kernel methods, computations involving $\phi$ are handled using kernel functions $k\left( {{x_i},{x_j}} \right) = \phi \left( {{x_i}} \right) \cdot \phi \left( {{x_j}} \right)$. In all of our experiments,  a Radial Basis Function (RBF) based kernel has been used. $C$ (fixed at 100) is the penalty parameter and $\zeta$ is the slack variable.

For multiple kernel learning (MKL), we follow the recipe by \cite{Sonnenburg2006} (\textit{cf.}  \cite{Xu2009}) and formulate a convex combination of sub-kernels as,

\begin{eqnarray}
{\mathbf{\kappa }}\left( {{x_i},{x_j}} \right) = \sum\limits_{k = 1}^K {{\beta _k}} {k_k}\left( {{x_i},{x_j}} \right)
\label{eqn:MKL}
\end{eqnarray}

In contrast to \cite{Sonnenburg2006}, we use L2 regularised ${\beta _k} \geqslant 0$ and $ \sum\limits_{k = 1}^K {{\beta _k}^{2} \leq 1} $. L2 regularised multiple kernel is learnt by formulating Eqn.\ref{eqn:MKL} as a semi-infinite linear programming (SILP) problem. During each iteration, a SVM solver\footnote{http://www.shogun-toolbox.org/} is first instantiated to obtain the weighted support vectors; subsequently, a linear programming (LP) problem is solved using Mosek\footnote{http://docs.mosek.com/6.0/toolbox/}.

\section{Results}

\begin{table}[]
\centering
\caption{SVM accuracy results for the Inception Network}
\label{table:inception}
\begin{tabular}{@{}llll@{}}
\toprule
        & optical flow & RGB    & MKL    \\ \midrule
split-1 & 61\%         & 54\%   & 68.1\% \\
split-2 & 62.4\%       & 50.8\% & 69.9\% \\
split-3 & 64\%         & 49.2\% & 70.0\% \\
Average & 62.5\%       & 51.3\% & 69.3\% \\ \bottomrule
\end{tabular}
\end{table}

\begin{table}[]
\centering
\caption{SVM accuracy results for the ResNet Network}
\label{table:resnet}
\begin{tabular}{@{}llll@{}}
\toprule
        & optical flow & RGB    & MKL    \\ \midrule
split-1 & 58.5\%       & 53.1\% & 64.1\% \\
split-2 & 57.5\%       & 48.6\% & 62.9\% \\
split-3 & 57.2\%       & 48\%   & 62.5\%   \\
Average & 57.7\%       & 49.9\% & 63.2\% \\ \bottomrule
\end{tabular}
\end{table}

\begin{table}[]
\centering
\caption{Fusing all kernels}
\label{table:average}
\begin{tabular}{@{}ll@{}}
\toprule
        & Accuracy \\ \midrule
split-1 & 73.1\%             \\
split-2 & 72.3\%             \\
split-3 & 72.9\%             \\
Average & 72.8\%             \\ \bottomrule
\end{tabular}
\end{table}

We use 3570 videos from the HMDB51 data-set for training the SVMs under a multiple kernel learning (MKL) framework. Utilising four networks yield four features tensors that are fused in steps, to form a single prediction (Figure \ref{fig:framework}). The feature tensors for both RGB and Flow are extracted from the output of the last connected layer with 1024 dimension for the Inception network and 2048 for the  ResNet network.  Four separate SVMs are trained on these feature tensors. Results have been shown for the two networks used -- Inception (Table \ref{table:inception}) and ResNet (Table \ref{table:resnet}). Subsequently, we fuse multiple kernels learnt from the individual classifiers using a semi-infinite linear optimisation problem.  Average result from three splits is displayed in Table \ref{table:average}. 

By fusing hand-crafted features, such as iDT and MIFS, with the features generated from a dCNN, the performance of pillar networks is further boosted. Such additional features take the place of `case-specific tensors' in Figure \ref{fig:framework}. It is apparent that combining kernels from various stages of the prediction process yields better accuracy.  Particularly, the confusion matrix suggests that the two worst performing classes are: (a) to wave one's hands, being confused with walking and (b) throwing objects confused with swinging a baseball. Unsurprisingly, actions such as chewing are confused with the eating action class. 

Table \ref{table:stateofart} compares our method to other methods in the literature. Of notable mention, are the TS-LSTM and the Temporal-Inception methods that form part of the framework that we use here. In short, synergistically, utilising multiple kernels boosts the performance of our classification framework, and enable state-of-the art performance on this dataset. The improvement using hand-crafted features i.e., iDT and MIFS are marginal. This means that the commonly seen boost of accuracy, offered by iDT is now being implicitly learnt by the combination of features learnt by the Inception and the ResNet pillars.

\begin{table*}[]
\centering
\caption{Accuracy scores for the HMDB51 data-set.}
\label{table:stateofart}
\begin{tabular}{@{}lll@{}}
\toprule
\textbf{Methods}   & \textbf{Accuracy {[}\%{]}} & \textbf{Reference} \\ \midrule
Two-stream         & 59.4                       &    \cite{Simonyan2014}                \\
Rank Pooling (ALL)+ HRP (CNN)         & 65                       &    \cite{Fernando2017}                \\
Convolutional Two-stream         & 65.4                       &  \cite{Feichtenhofer2016}                  \\
Temporal-Inception & 67.5                       &  \cite{Ma2017}                  \\
TS-LSTM            & 69                         & \cite{Ma2017}                   \\
Temporal Segment Network (2/3/7 modalities) & 68.5/69.4/71                       &  \cite{Wang2016}                  \\
ST-ResNet + iDT            & 70.3                         & \cite{Ma2017}                   \\
ST-multiplier network            & 68.9                         & \cite{Feichtenhofer2017}                    \\
ST-multiplier network + iDT    &     72.2                        & \cite{Feichtenhofer2017}
\\
Pillar Networks + SVM-MKL    &     72.8                        & this paper                   \\ 
Pillar Networks + iDT + SVM-MKL &     73.0                        & this paper                  
\\
Pillar Networks + MIFS + SVM-MKL    &     73.3                        & this paper                   \\
\bottomrule
\end{tabular}
\end{table*}

\section{Discussion}

Our main contribution in this paper is to introduce \textbf{pillar networks} that are deep as well as wide (by plugging in other feature tensors, horizontally) enabling horizontal scalability. Combining different methodologies allow us to reach state-of-the-art performance for video classification, especially, action recognition.

We utilised the HMDB-51 dataset instead of UCF101 as the former has proven to be difficult for deep networks due to the heterogeneity of image quality, camera angles, etc. As is well-known videos contain extensive long-range temporal structure; using different networks (2 ResNets and 2 Inception networks) to capture the subtleties of this temporal structure is an absolute requirement. Since each network implements a  different non-linear transformation, one can utilise them to learn very deep features. Utilising the distributed architecture then enables us to parcellate the feature tensors into computable chunks (by being distributed) of input for an SVM-MKL classifier. Such an architectural choice, therefore, enables us to scale horizontally by plugging in a variety of networks \textit{as per requirement}. While we have used this architecture for video based classification, there is a wide variety of problems where we can apply this methodology -- from speech processing (with different pillars/networks) to natural-language-processing (NLP).

In supplementary studies, stacking features from the four network outputs with a softmax and a cross-entropy loss has an accuracy of approximately 67\%, highlighting the necessity for a multi-kernel approach. Thus, our framework rests on two stages of training -- one for training the neural networks and the other for training the multiple kernels of the support vector machine (SVM). Since both of the training stages are decoupled, it allows for scalability wherein different networks can operate on a plug-and-play basis.  Indeed, there has been some work in combining deep neural networks with (linear) SVMs to facilitate end-to-end training \cite{Tang2013}. 

It would be useful to see how pillar networks perform on immensely large datasets such as the Youtube-8m data-set \cite{Abu-El-Haija2016}. Additionally, recently published  Kinetics human action video dataset from DeepMind \cite{Kay2017} is equally attractive, as pre-training, the pillar networks on this dataset before fine-grained training on HMDB-51 will invariably increase the accuracy of the current network architecture. 

 
{\small
\bibliographystyle{ieee}
\bibliography{video_action}
}

\end{document}